%% file: mainTemp.tex
\documentclass{article}

\usepackage{arxiv}

\usepackage[utf8]{inputenc} 
\usepackage[T1]{fontenc}    
\usepackage{hyperref}       
\usepackage{url}            
\usepackage{booktabs}       
\usepackage{amsfonts}       
\usepackage{nicefrac}       
\usepackage{microtype}      
\usepackage{lipsum}

\usepackage{graphicx}
\usepackage{balance}  
\usepackage{lipsum}
\usepackage{tabularx}
\usepackage{graphicx}
\usepackage{epstopdf}
\usepackage{psfrag}
\usepackage{color}
\usepackage{filecontents}
\usepackage{cite}
\usepackage{mathtools}

\usepackage{amssymb,amsmath,amsfonts}

\usepackage{float}
\usepackage{algorithm}
\usepackage[noend]{algpseudocode}

\usepackage{tikz}
\usepackage{ifthen}
\usepackage{subfigure}
\usepackage{overpic}
\usepackage{pict2e}

\usepackage{caption}
\usepackage{subfigure,float}
\usepackage{footnote}

\newcommand{\HMatrix}{\mathbf{H}}
\newcommand{\XMatrix}{\mathbf{X}}
\newcommand{\YMatrix}{\mathbf{Y}}
\newcommand{\AMatrix}{\mathbf{A}}
\newcommand{\IMatrix}{\mathbf{I}}

\newcommand{\ZMatrix}{\mathbf{Z}}
\newcommand{\DMatrix}{\mathbf{D}}

\newcommand{\network}{\cal{G}}

\newcommand{\nodeset}{\cal{V}}
\newcommand{\edgeset}{\cal{E}}

\title{Graph Convolutional Networks Meet with High Dimensionality Reduction}

\author{
  Mustafa Co{\c s}kun \\
  Department of Computer Engineering\\
  Abdullah  G{\"ul} University\\
  Kayseri, Turkey 38100 \\
  \texttt{mustafa.coskun@agu.edu.tr} \\
}

\begin{document}
\maketitle

\input{abstract}

\keywords{First keyword \and Second keyword \and More}

\input{introduction}
\input{methods}
\input{Experiments}
\input{Conclusion}

\section*{Acknowledgment}
We would like to thank Professor Dr. Mehmet Koyut{\"u}rk from department of Data \& Computer Science at Case Western Reserve University, USA for his many insightful discussions on this work.  
\bibliographystyle{smppsci}      
\bibliography{mybib} 




\end{document}

%% file: abstract.tex
\begin{abstract}Recently, Graph Convolutional Networks (GCNs) and their variants have been receiving many research interests for learning graph-related tasks. These tasks include, but not limited to, link prediction, node classification, among many others. In the node classification problem, the input is a graph with some labeled nodes, edges and features associated with these nodes and the objective is to predict the labels of the nodes that are not labeled, using graph topology as well as the features of the labeled nodes. While the GCNs have been successfully applied to this problem, some caveats inherited from classical deep learning still remain as open research topics in the context of the node classification problem. One such inherited caveat is that GCNs only consider the nodes that are a few propagations away from the labeled nodes to classify them. However, taking only a few propagation steps away nodes into account defeats the purpose of using the graph topological information in the GCNs. To remedy this problem, the-state-of-the-art methods leverage the network diffusion approaches, namely personalized page rank and its variants, to fully account for the graph topology, {\em after} they use the Neural Networks in the GCNs. However, these approaches overlook the fact that the network diffusion methods favour high degree nodes in the graph, resulting in the propagation of labels to unlabeled centralized, hub, nodes. To address this biasing hub nodes problem, in this paper, we propose to utilize a dimensionality reduction technique conjugate with personalized page rank so that we can both take advantage from graph topology and resolve the hub node favouring problem for GCNs. Here, our approach opens a new holistic road for message passing phase of GCNs by suggesting usage other proximity matrices instead of well-known Laplacian. Testing on two real-world networks that are commonly used in benchmarking GCNs' performance for the node classification context, we systematically evaluate the performance of the proposed methodology and show that our approach outperforms existing methods for wide ranges of parameter values with very limited deep learning training {\em epochs}. \href{https://github.com/mustafaCoskunAgu/ScNP/blob/master/TRJMain.m}{The code} is freely available. 
\end{abstract}

%% file: introduction.tex
\section{Introduction}
\label{sec:introduction}
Graph Convolutional Networks (GCNs)~\cite{Kipf} are a versions of Convolutional Neural Netwoks (CNNs) on graphs\cite{FWu}. To learn the graph representations, GCNs utilize layers of learned filters followed by a nonlinear activation function~\cite{FWu}. In recent years,  GCNs have been successfully applied to wide range of problems in data mining, including node classification~\cite{Kipf}, recommendation systems~\cite{YingKdd}, the prediction of combined side effects of drugs (polypharmacy side effects) ~\cite{zitnik2018modeling}, and natural language processing~\cite{yao2019graph}.

In node classification problem, GCNs take a graph, which represents the relationship among vertices/nodes via edges connecting them, features associated with the vertices and label information of some --not all-- nodes as an input. The objective of GCNs is then to predict the labels of rest of the vertices in the graph, using the features as well as the graph topology based on the rationale that neighboring nodes should exhibit similar labels~\cite{AAAI18Li}. To achieve the node classification aim, at each layer of GCNs, the convolution is performed by applying a first-order spectral filter to the feature matrix, followed by a nonlinear activation function~\cite{FWu}. Thereby, the features are smoothed across the graph at each layer of the neural network by using graph's connectivity, and this process is also known as the message passing phase of a GCN~\cite{Kipf}.

Despite the successful application of GCNs to a plethora important problems, one subtle issue, which is an inherited caveat from traditional deep learning, remains unresolved: as in many other deep learning applications, the message passing scheme of GCNs only utilizes a few hop neighborhoods of labeled nodes~\cite{klicpera2018combining}. There has been a few recent attempts to address the limited message passing scheme of GCNs by using attention mechanisms~\cite{hamilton2017inductive}, random walk~\cite{abu2018watch}, and edge features~\cite{schlichtkrull2018modeling}.  However, all of these methods can only utilize the graph topology up to very few neighbors for each node~\cite{klicpera2018combining} since using many convolution layers in the message passing of GCNs could potentially be detrimental to original classification task and mix the predicted labels via over-smoothing of the features~\cite{AAAI18Li}. This over-smoothing is the motivated/forcing primary reason using two layers GCNs instead of many layered GCNs, which is a more intuitive approach than using two layers GCNs to propagate the labels to the far away nodes in the graph~\cite{AAAI18Li, Kipf}.

To circumvent the limited message passing problem of GCNs, Klicpera et al.~\cite{klicpera2018combining} first observe the connection between feature propagation in GCNs and well-known PareRank algorithm~\cite{PageRank}. Then, they propose an algorithmic framework that utilizes {\em personalized} version of PageRank rather than using message passing phase of GCNs to nicely segregate the propagation scheme from neural networks~\cite{klicpera2018combining}.  
In turn, this type of propagation scheme permits the use of far more (in fact, infinitely many) propagation steps without leading to over-smoothing problem as in the propagation scheme of the GCNs' feature propagation phase, i.e., message passing scheme ~\cite{klicpera2018combining}. While this approach has been shown to be effective in rendering the application of GCNs to instances with many layers neural networks without over-smoothing problem, it ignores an important problem that is associated with the application of {\em personalized} PageRank based techniques: The random walker in {\em personalized} PageRank based assessment of network proximity assigns higher scores to nodes with high connectivity and/or centrality~\cite{dada, CoskunLink}, thus biasing the scoring toward highly connected nodes.

Inspired by our earlier work in the context of link prediction ~\cite{CoskunLink}, here we propose an algorithmic framework that  fairly assesses the similarity of the nodes in a graph, i.e., without being forced by the high degree connectivity of individual nodes as in {\em personalized} PageRank~\cite{klicpera2018combining} . More specifically, our approach is based on the idea that nodes that are topologically ``similar?--that is, if their corresponding columns in fully computed {\em personalized} PageRank matrix are ``similar" as whole vectors in terms of any vector comparison metric, such as Pearson correlation, then they should be ``similar" to each other. In other words, as opposed to directly assessing the proximity of two nodes by relying on entries in the columns of full {\em personalized} PageRank matrix, we measure their similarity in terms of what they are close to. As we have shown previously in the context link prediction~\cite{CoskunLink}, this approach drastically reduces degree bias in measuring closeness of the nodes in {\em personalized} PageRank matrix. While assessing the proximity based on topological profile closeness is useful, sparseness of the underlying network might adversely affects the performance of this approach due to high dimensionality problem~\cite{CoskunLink}. Rather, here in this paper, we first reduce the high-dimensionality in the columns of {\em personalized} PageRank matrix by just discarding the entries which are smaller than a given parameter, $\epsilon$, when we measure the closeness of two nodes and account for the correlation of those reduced profile vectors. Finally, in the context of node classification problem in GCN, we assess the reduced topological profile similarity matrix rather than {\em personalized} PageRank matrix in~\cite{klicpera2018combining}.

To test the performance of the proposed methodology in improving the accuracy of GCN-based node classification and reducing the train time via less training {\em epochs},which refers to one cycle through the full training dataset, we perform systematical experiments on two citation networks.
Our results show that the proposed approach using reduced topological profile similarity renders
GCNs highly effective in node classification, and the resulting algorithmic framework outperform algorithm using {\em personalized} PageRank~\cite{klicpera2018combining} with very limited {\em epoch} numbers.

The rest of the paper is organized as follows: in Section~\ref{sec:method}, we describe the terminology, establish background on GCNs and {\em personalized} PageRank~\cite{klicpera2018combining} approaches instead of message passing phase in classical GCNs, and describe our approach.In Section~\ref{sec:experiments}, we provide detailed experimental evaluation of our proposed approach. We draw conclusions and summarize avenues for further research in Section~\ref{sec:conclusion}.

%% file: methods.tex
\section{Methods}
\label{sec:method}

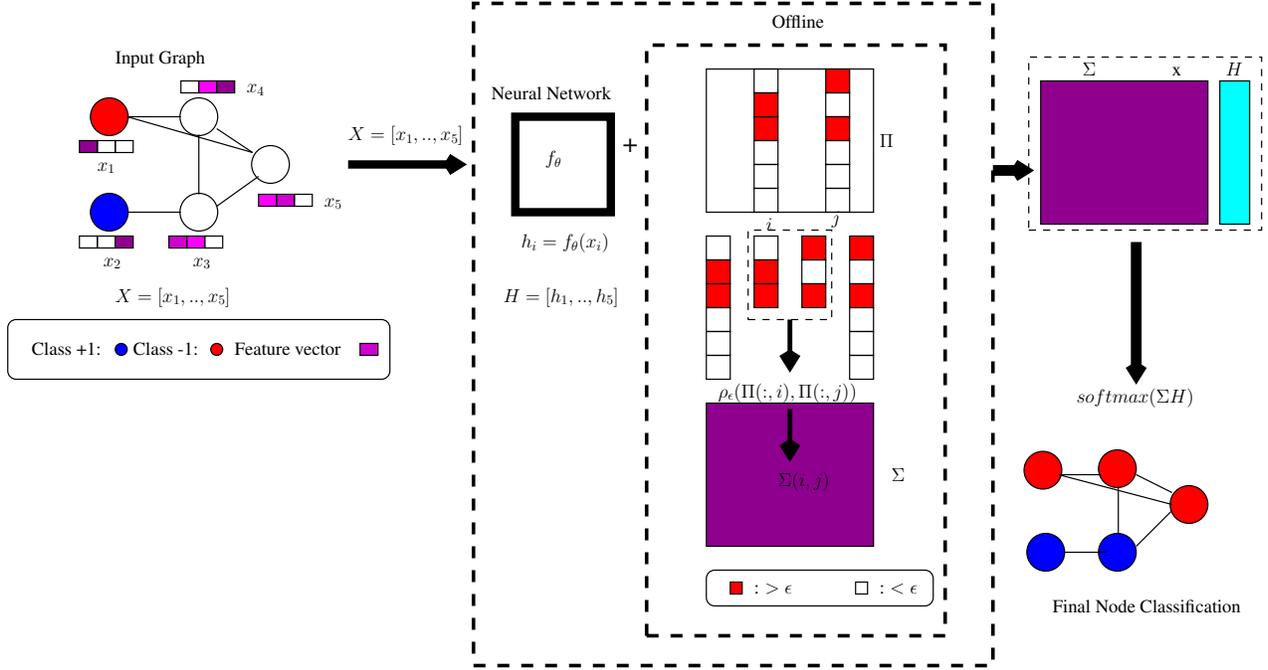
\begin{figure*}[t]
	\centering
	\scalebox{0.5}{\input{MainFigTRJ.pspdftex}}
	\caption{\textbf{Flowchart illustrating the proposed approach , for separating a Graph Convolutional Network (GCN)'s Neural Network phase from its propagation phase.} Given an undirected graphs with features associated with all nodes and labels associated with some nodes (red and blue nodes in the first graph),  performs the following steps: (i)  constructing $\HMatrix \in \mathbb{R}^{n\times c}$ matrix via the training Neural Network with on feature information of the nodes, (ii) offline computation of $\Pi$ matrix to obtain a low-dimensional topological profile (by eliminating the entries of topological profiles of node $i$ and $j$ if they both have common white box (meaning entries less than given $\epsilon$ for both of them; otherwise we keep the rows )) for each node and constructing $\Sigma$, {\em Sparse Correlation Matrix} , (iii) . Finally, we use $\Sigma$ and $\HMatrix$ to further refine the classification of nodes.} 
\label{fig:mainFigure}
\end{figure*}
In this section, we first define graph convolutional networks(GCNs) in the context of node classification problem. We then present insights for the usage of {\em personalized} PageRank to separate the message passing scheme of GCNs from neural networks. Subsequently, we show that considering global correlation of nodes can be integrated into GCNs massage passing framework to balance the propagation toward intrinsic connectivity.
\subsection{Graph Convolutional Networks}
Here, we follow the notations by Kipf and Welling~\cite{Kipf} to define GCNs in the context of the node classification problem. Let $\network = (\nodeset, \edgeset)$ be an undirected graph, where $\nodeset$ denotes set of $n$ nodes and $\edgeset$ represents the set of $m$ edges in the graph. The nodes are associated with a feature matrix ${\XMatrix} = {[x_1, x_2, ... ,x_n]} \in \mathbb{R}^{n \times f}$ such that $x_i \in \mathbb{R}^{f}$ is a feature vector for node $v_i \in {\nodeset}$ and the label matrix is given as $\YMatrix \in \mathbb{R}^{n \times c}$, with number of classes $c$. The adjacency matrix of $\network$ is given by $\AMatrix \in \mathbb{R}^{n\times n}$, where $\AMatrix$'s entries are the weights of edges connecting nodes in $\network$. Let $\tilde{\AMatrix} : \AMatrix + \IMatrix_n$ be self-loop added adjacency matrix of $\network$. Then, we can give node classification problem as follows:\\
\textbf{Problem (Node Classification):} We are given an undirected graph and features of the nodes in this graph and a set of labeled nodes and our aim is to predict the rest of labels of all nodes in the graph.

One simple yet very effective message passing algorithm that is used aforementioned problem is the GCN. Now, we can give the two message passing layers of GCNs as follows~\cite{Kipf}

\begin{equation}
\ZMatrix_GCN = softmax(\hat{\tilde{\AMatrix}}ReLU(\hat{\tilde{\AMatrix}}\XMatrix \Theta^{(0)} ) \Theta^{(1)}),
\label{eq:TwoGCN}    
\end{equation}

where $\ZMatrix \in \mathbb{R}^{n \times c}$ are the predicted labels, $\hat{\tilde{\AMatrix}} = \tilde{D}^{-1/2}\tilde{A} \tilde{D}^{-1/2}$ is symmetrically normalized adjacency matrix, $\tilde{A}$, $\tilde{\DMatrix}$ is diagonal matrix with degrees of $\tilde{A}$, and $\Theta^{(0)} )$ and $ \Theta^{(1)}$ are trainable weight matrices~\cite{Kipf}.

For the node classification problem, we employ equation~(\ref{eq:TwoGCN}) to predict the labels of all nodes that are not labeled by only considering two-hop away propagation. In essence, there are mainly two reasons why we cannot use a larger propagation steps in GCNs: first, using many layers are equivalent to Laplacian smoothing~\cite{AAAI18Li} and second increasing dept of neural networks causes increasing and complicating the GCNs~\cite{klicpera2018combining}. However, dept of neural networks and usage of many steps in propagation two complementary aspects to each other. The Usage of the shallow layered networks leads to bad compromises~\cite{klicpera2018combining}.

Clearly, using many propagation steps is crucial to determine the similar labeled nodes which are reside in the different/distinct regions of the graph, however, we cannot employ many graph convolutional layers due to over-soothing problem highlighted by~\cite{AAAI18Li}. To demystify this problem, Xu et al.~\cite{AAAI18Li} have shown that for a $k$-layer GCN, the influence score of node $x$ on node $y$ exhibits random walk alike properties on a graph. More specifically, they observe that influence score of $x$ on $y$ can be obtained by solving the first eigenvalue problem as $\pi^{*} = \hat{\tilde{\AMatrix}} \pi^{*}$, where $\pi^{*}$ denotes converged first eigenvector~\cite{AAAI18Li}. Obviously, this propagation relies more on graph rather than node where the random walker has started to walk and its neighborhood. To alleviate this problem, Klicpera et al.~\cite{klicpera2018combining} exploits random walk restarts(RWR) approach to have the random walker return the start node once a while. We give brief overview of the idea that used by Klicpera et al.~\cite{klicpera2018combining} to integrate RWR into message passing framework in the following subsection.

\subsection{Existing Solution to Message Passing via Personalized Propagation of Neural Predictions}
\textbf{From message passing to RWR:} Original PageRank algorithm has been used in numerous applications~\cite{PageRank} and it is computed by only solving $\pi_{pr} = \AMatrix_{rw} \pi_pr$ the first eigenvalue problem, where $\AMatrix_{rw} = \AMatrix \DMatrix^{-1}$, row-normalized adjacency matrix. To account for the influence score of the root node, starting nodes and its neighborhood, Klicpera et al.~\cite{klicpera2018combining} use the personalized variant of PageRank~\cite{PageRank}, i.e, Random Walk with Restarts, as follows:

\begin{equation}
\pi_{ppr}(i_x) = \alpha {(\IMatrix_n - (1-\alpha)\hat{\tilde{\AMatrix}})}^{-1} i_x
\label{eq:RWR}    
\end{equation}
where $\alpha \in (0,1]$ is the teleport probability that determines how much a random walker should span the graph and $i_x$ indicator vector which only contains $1$ its $x$-th entries and $0$s in all other entries. Clearly, $\Pi_{ppr} = \alpha {(\IMatrix_n - (1-\alpha)\hat{\tilde{\AMatrix}})}^{-1}$ is an $n\times n$ fully personalized PageRank matrix that contains influence score of $x$ on $y$, its $\Pi_{ppr}(x,y) = \Pi_{ppr}(y,x)$ entry.\\
\textbf{Personalized propagation of neural predictions (PPNP)~\cite{klicpera2018combining}:} In a nutshell, to utilize the fully personalized PageRank scores for node classification problem, Klicpera et al.~\cite{klicpera2018combining} first generate classification prediction of each node based on its features and then propagate these prediction to gain more confidence about final prediction via fully personalized PageRank~\cite{klicpera2018combining}. More formally, this formulation of node classification can be given as follows~\cite{klicpera2018combining}: 

\begin{equation}
\ZMatrix_{PPNP} = softmax(\Pi_{ppr} \HMatrix)
\label{eq:PPNP}    
\end{equation}
where $\HMatrix = f_{\theta}(\XMatrix)$ and $\XMatrix$ denotes features matrix and $f_{\theta}$ a neural network with parameter set $\theta$ generating the first predictions, $\HMatrix \in \mathbb{R}^{n \times c}$ before applying fully personalized PageRank and $softmax(x) = \dfrac{1}{\sum_{j=1}^{k} exp(x_c)}exp(x) $ transforms
predicted values into the probability density function~\cite{FWu}.

As a result, PPNP~\cite{klicpera2018combining} separates message passing of GCNs from the neural networks that is used for generating prediction scores~\cite{klicpera2018combining}. Now, with the help of this formulation, the depth of neural networks are fully separated from propagation phase of GCNs. 

Furthermore, Klicpera et al.~\cite{klicpera2018combining} propose to compute equation~(\ref{eq:RWR}) via classical power iteration to give less accurate but more efficient variant of the propagation step as follows:
\begin{equation}
\begin{aligned}
\ZMatrix^{(0)} & =
\HMatrix = f_{\theta}(X),
\\
\ZMatrix^{(k+1)}& = (1-\alpha)\hat{\tilde{\AMatrix}}\ZMatrix^{(k)} + \alpha \HMatrix,  
\\
\ZMatrix^{(K)}& = softmax((1-\alpha)\hat{\tilde{\AMatrix}}\ZMatrix^{(K-1)} + \alpha \HMatrix).
\end{aligned}
\label{eq:appnr}
\end{equation}
where $k \in [0,K-2]$, dimension of power iteration.
\subsection{Proposed Solution to the Message Passing Scheme.} 
The main idea behind the proposed approach is that  that the proximity of nodes, influence score of node $x$ on $y$,  in a graph can be exploited more effectively by considering the {\em relative} position of nodes in the graph with respect to each other. 
In other words, if we consider the closeness of two nodes from perspective of similar to nodes, we might gain more insights in about segregated propagation steps, i.e, personalized PageRank. 
In the context of link prediction~\cite{CoskunLink} and its applications and drug response prediction~\cite{CoskunDrug}, we demonstrate shown that this approach is indeed more effective than personalized PageRank like proximity measurements, resulting much more accurate prediction results by drastically eliminating high dimensionality of PageRank vectos , such as connectivity and/or centrality~\cite{vavien}.
Elimination of such bias is particularly important for the node classification problem, since final predictions proposed in~\cite{klicpera2018combining} can be biased towards to the high degree nodes and misguides the entire prediction process by favouring central unlabeled nodes.

Mathematically, it is clear from the equation~(\ref{eq:RWR}) that the influence of a node $x$ on all the other nodes, including itself, can be given the vector $\pi_{ppr}(i_x)$. Now, considering all nodes influence on all other nodes in the graph is given by the matrix, $\Pi_{ppr} I_n=  \alpha {(\IMatrix_n - (1-\alpha)\hat{\tilde{\AMatrix}})}^{-1} I_n$. Then, Klicpera et al~\cite{klicpera2018combining} propose their \textbf{PPNP} approach based on $\alpha {(\IMatrix_n - (1-\alpha)\hat{\tilde{\AMatrix}})}^{-1}$ matrix's entries distribution by considering individual entries' influences on themselves. Rather, here, we propose to account for $\pi_{ppr}(i_x)$ vector as whole to measure the influences of two nodes each other. More formally, for a given  $\Pi_{ppr} = \alpha {(\IMatrix_n - (1-\alpha)\hat{\tilde{\AMatrix}})}^{-1}$ fully computed personalized PageRank matrix, we rely on the proximity of correlations of two nodes $x$ and $y$ as $\rho (\Pi_{ppr}(:, x), \Pi_{ppr}(:,y))$, where $\rho$ denotes Pearson correlation of two column vectors corresponding profile vectors of nodes $x$ and $y$, respectively. This way, we capture the global profile information of nodes from the perspective of all nodes in the graph by jointly considering that if a random walker starts walking from a specific nodes,$x$, and what would be probabilities of landing all other nodes. Then, we take the closeness measure in terms of these landing probabilities when we evaluate the proximity of two nodes, $x$ and $y$. This approach has been shown to be much more informative than considering individual proximity separately  in the context of disease gene prioritization~\cite{dada, vavien} when one wants to classify nodes' belongings. 

However, the high-dimensionality problem of vector $\Pi_{ppr}(:, x)$ for any nodes $x$ in graph may prevent us applying Pearson correlation to those profile vectors~\cite{CoskunLink,CoskunDrug} due to sparsity of the vectors. Or, unveil the hints to develop more efficient algorithms in terms of computational and storage cost.

To this end, we propose to reduce the high-dimensionality of all vectors by pruning the the rows of the any two vectors that are less than given parameter $\epsilon$. That is, we delete the rows of  $\Pi_{ppr}(:, x)$ and $\Pi_{ppr}(:, y)$, if they simultaneously smaller than the $\epsilon$ to prevent small entries of both vectors act as if they are correlated. This technique is named as {\em Sparse Correlation}, $\rho_{\epsilon}$, in~\cite{CoskunLink}, to be consisted with the pioneering paper~\cite{CoskunLink}, we use the same terminology for the rest of this paper.

Now, our objective should be clear that we aim at using sparsely correlated columns of $\Pi_{ppr}$ in the context of node classification via GCN. More precisely, we construct a {\em Sparse Correlation Matrix}, $\Sigma$ from $\Pi_{ppr}(:, x)$ and provide it to any GCNs to classify the nodes. The construction of this $\Sigma$ matrix is given below algorithm:

\begin{algorithm}
\caption{\textbf{Offline} Construction of $\Sigma$ }\label{alg:HeuristicAlgo}
\begin{algorithmic}[1]
\State Compute $\Pi_{ppr} \in \mathbb{R}^{n \times n}$ 
 \For{\texttt{i=1:n}}
    \For{\texttt{j=1:i}}
      \State $\Sigma(i,j) = \rho_{\epsilon} (\Pi_{ppr}(:,i), \Pi_{ppr}(:,j))$
    
    \EndFor     
 \EndFor
 \State $\Sigma = \Sigma + \Sigma^{T}$ \Comment{$(.)^{T}$: Transpose of a matrix}
\end{algorithmic}
\end{algorithm}

Upon the computation of $\Sigma$ matrix, we give following variant, {\sc Sparse correlation of Neural Predictions (ScNP) }, of GCN algorithm for node classification problem:
\begin{equation}
\ZMatrix_{\sc ScNP} = softmax(\Sigma \HMatrix)
\label{eq:SCNP}    
\end{equation}

In the final step of {\sc ScNP}, we use \textit{softmax} classifier to classify nodes based on their learned features via neaural networks and {\em Sparse Correlation Matrix}, $\Sigma$. With the proposed approach, we take the landmarks, which can be seen as the views of closely related nodes for the nodes that we want to classify, as oppose to a single entry view point as in~\cite{klicpera2018combining}. The flowchart of our approach can be seen in Figure~\ref{fig:mainFigure}. 

%% file: MainFigTRJ.pspdftex
\begin{picture}(0,0)%
\includegraphics{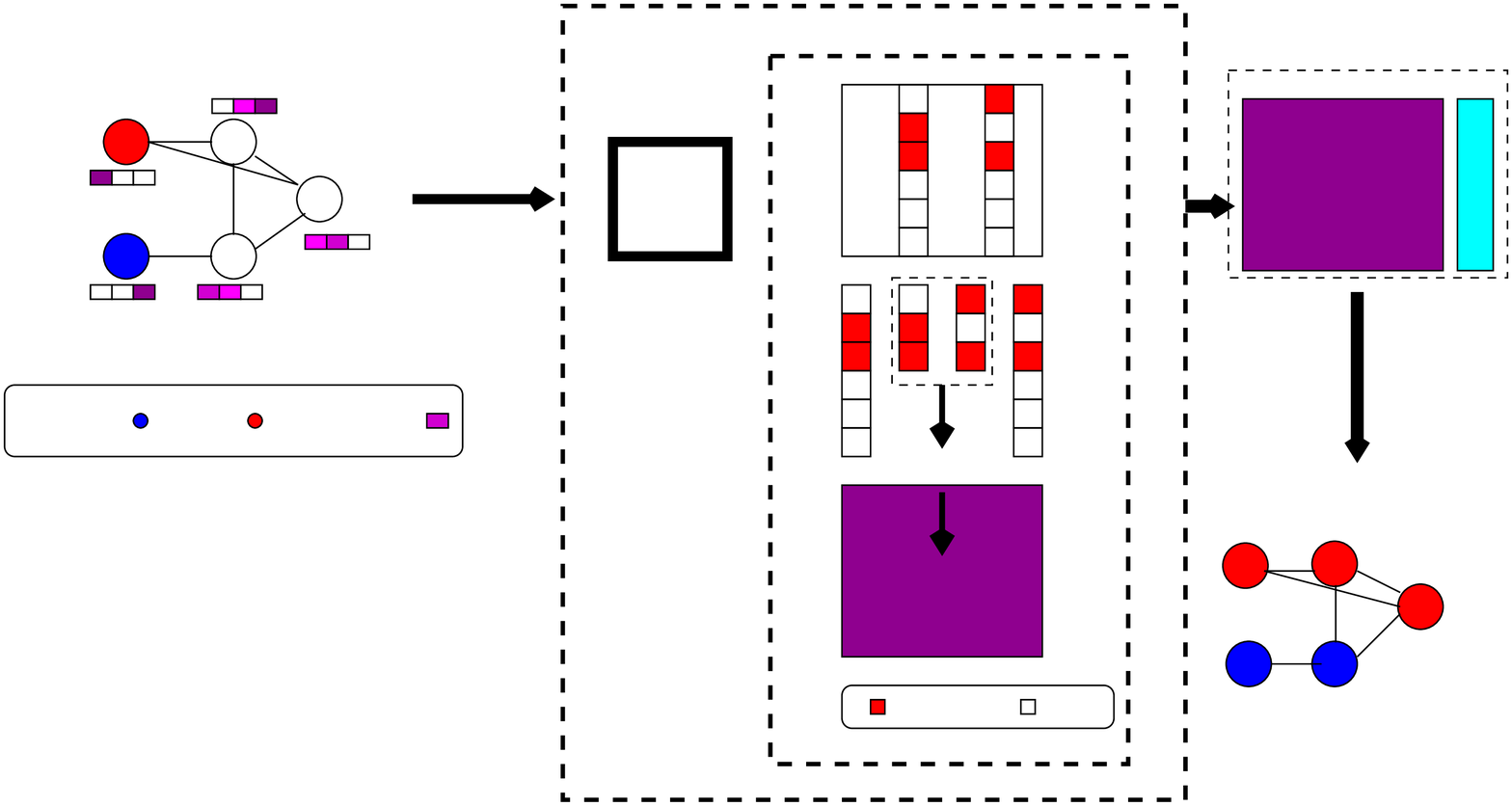}%
\end{picture}%
\setlength{\unitlength}{3947sp}%
\begingroup\makeatletter\ifx\SetFigFont\undefined%
\gdef\SetFigFont#1#2#3#4#5{%
  \reset@font\fontsize{#1}{#2pt}%
  \fontfamily{#3}\fontseries{#4}\fontshape{#5}%
  \selectfont}%
\fi\endgroup%
\begin{picture}(15794,8413)(1104,-7905)
\put(10051,-4486){\makebox(0,0)[lb]{\smash{{\SetFigFont{14}{16.8}{\rmdefault}{\mddefault}{\updefault}{\color[rgb]{0,0,0}$\rho_{\epsilon}(\Pi(:,i), \Pi(:,j))$}%
}}}}
\put(2476,-286){\makebox(0,0)[lb]{\smash{{\SetFigFont{14}{16.8}{\rmdefault}{\mddefault}{\updefault}{\color[rgb]{0,0,0}Input Graph}%
}}}}
\put(2251,-1636){\makebox(0,0)[lb]{\smash{{\SetFigFont{14}{16.8}{\rmdefault}{\mddefault}{\updefault}{\color[rgb]{0,0,0}$x_1$}%
}}}}
\put(2326,-2836){\makebox(0,0)[lb]{\smash{{\SetFigFont{14}{16.8}{\rmdefault}{\mddefault}{\updefault}{\color[rgb]{0,0,0}$x_2$}%
}}}}
\put(3451,-2836){\makebox(0,0)[lb]{\smash{{\SetFigFont{14}{16.8}{\rmdefault}{\mddefault}{\updefault}{\color[rgb]{0,0,0}$x_3$}%
}}}}
\put(5101,-2086){\makebox(0,0)[lb]{\smash{{\SetFigFont{14}{16.8}{\rmdefault}{\mddefault}{\updefault}{\color[rgb]{0,0,0}$x_5$}%
}}}}
\put(4126,-661){\makebox(0,0)[lb]{\smash{{\SetFigFont{14}{16.8}{\rmdefault}{\mddefault}{\updefault}{\color[rgb]{0,0,0}$x_4$}%
}}}}
\put(2476,-3286){\makebox(0,0)[lb]{\smash{{\SetFigFont{14}{16.8}{\rmdefault}{\mddefault}{\updefault}{\color[rgb]{0,0,0}${X} = [x_1,..,x_5]$}%
}}}}
\put(1426,-3961){\makebox(0,0)[lb]{\smash{{\SetFigFont{14}{16.8}{\rmdefault}{\mddefault}{\updefault}{\color[rgb]{0,0,0}Class +1: }%
}}}}
\put(2701,-3961){\makebox(0,0)[lb]{\smash{{\SetFigFont{14}{16.8}{\rmdefault}{\mddefault}{\updefault}{\color[rgb]{0,0,0}Class -1: }%
}}}}
\put(3976,-3961){\makebox(0,0)[lb]{\smash{{\SetFigFont{14}{16.8}{\rmdefault}{\mddefault}{\updefault}{\color[rgb]{0,0,0}Feature vector }%
}}}}
\put(5401,-1261){\makebox(0,0)[lb]{\smash{{\SetFigFont{14}{16.8}{\rmdefault}{\mddefault}{\updefault}{\color[rgb]{0,0,0}${X} = [x_1,..,x_5]$}%
}}}}
\put(7201,-736){\makebox(0,0)[lb]{\smash{{\SetFigFont{14}{16.8}{\rmdefault}{\mddefault}{\updefault}{\color[rgb]{0,0,0}Neural Network}%
}}}}
\put(7576,-2611){\makebox(0,0)[lb]{\smash{{\SetFigFont{14}{16.8}{\rmdefault}{\mddefault}{\updefault}{\color[rgb]{0,0,0}$h_i= f_{\theta}(x_i)$ }%
}}}}
\put(7351,-3286){\makebox(0,0)[lb]{\smash{{\SetFigFont{14}{16.8}{\rmdefault}{\mddefault}{\updefault}{\color[rgb]{0,0,0}${H} = [h_1,..,h_5]$}%
}}}}
\put(10801,-5611){\makebox(0,0)[lb]{\smash{{\SetFigFont{14}{16.8}{\rmdefault}{\mddefault}{\updefault}{\color[rgb]{0,0,0}$\Sigma(i,j)$}%
}}}}
\put(10651,-2386){\makebox(0,0)[lb]{\smash{{\SetFigFont{14}{16.8}{\rmdefault}{\mddefault}{\updefault}{\color[rgb]{0,0,0}$i$}%
}}}}
\put(11476,-2311){\makebox(0,0)[lb]{\smash{{\SetFigFont{14}{16.8}{\rmdefault}{\mddefault}{\updefault}{\color[rgb]{0,0,0}$j$}%
}}}}
\put(10726,164){\makebox(0,0)[lb]{\smash{{\SetFigFont{14}{16.8}{\rmdefault}{\mddefault}{\updefault}{\color[rgb]{0,0,0}Offline}%
}}}}
\put(10501,-6961){\makebox(0,0)[lb]{\smash{{\SetFigFont{14}{16.8}{\rmdefault}{\mddefault}{\updefault}{\color[rgb]{0,0,0}: $>\epsilon$ }%
}}}}
\put(12076,-6961){\makebox(0,0)[lb]{\smash{{\SetFigFont{14}{16.8}{\rmdefault}{\mddefault}{\updefault}{\color[rgb]{0,0,0}: $<\epsilon$}%
}}}}
\put(12076,-1336){\makebox(0,0)[lb]{\smash{{\SetFigFont{14}{16.8}{\rmdefault}{\mddefault}{\updefault}{\color[rgb]{0,0,0}$\Pi$}%
}}}}
\put(12226,-5536){\makebox(0,0)[lb]{\smash{{\SetFigFont{14}{16.8}{\rmdefault}{\mddefault}{\updefault}{\color[rgb]{0,0,0}$\Sigma$}%
}}}}
\put(8851,-1411){\makebox(0,0)[lb]{\smash{{\SetFigFont{20}{24.0}{\rmdefault}{\mddefault}{\updefault}{\color[rgb]{0,0,0}+}%
}}}}
\put(14626,-436){\makebox(0,0)[lb]{\smash{{\SetFigFont{14}{16.8}{\rmdefault}{\mddefault}{\updefault}{\color[rgb]{0,0,0}$\Sigma$}%
}}}}
\put(16426,-436){\makebox(0,0)[lb]{\smash{{\SetFigFont{14}{16.8}{\rmdefault}{\mddefault}{\updefault}{\color[rgb]{0,0,0}$H$}%
}}}}
\put(14551,-4561){\makebox(0,0)[lb]{\smash{{\SetFigFont{14}{16.8}{\rmdefault}{\mddefault}{\updefault}{\color[rgb]{0,0,0}$softmax(\Sigma H)$}%
}}}}
\put(15751,-436){\makebox(0,0)[lb]{\smash{{\SetFigFont{14}{16.8}{\rmdefault}{\mddefault}{\updefault}{\color[rgb]{0,0,0}x}%
}}}}
\put(14251,-7186){\makebox(0,0)[lb]{\smash{{\SetFigFont{14}{16.8}{\rmdefault}{\mddefault}{\updefault}{\color[rgb]{0,0,0}Final Node Classification}%
}}}}
\put(7876,-1561){\makebox(0,0)[lb]{\smash{{\SetFigFont{14}{16.8}{\rmdefault}{\mddefault}{\updefault}{\color[rgb]{0,0,0}$f_{\theta}$}%
}}}}
\end{picture}%

%% file: Experiments.tex
\section{Experiments}
\label{sec:experiments}
\begin{table}[t]
\caption{Descriptive Statics of Datasets}
\label{datatable}
\centering
\resizebox{0.55\textwidth}{!}{%
\begin{tabular}{|c| c c c c c|}
\hline
Network &  Type & Nodes &  Edges &  Classes &  Features  \\
\hline\hline
 {\tt Cora-ML} & Citation & 2810 &  7981 &  7 & 2879  \\
 {\tt CiteSeer} &Citation & 2110  &  3668 & 6  & 3703  \\
\hline
\end{tabular}
}
\end{table}

\begin{figure}[htbp]
  \begin{center}
  \subfigure[]{\label{fig:Result2a}
     \includegraphics[scale=0.5]{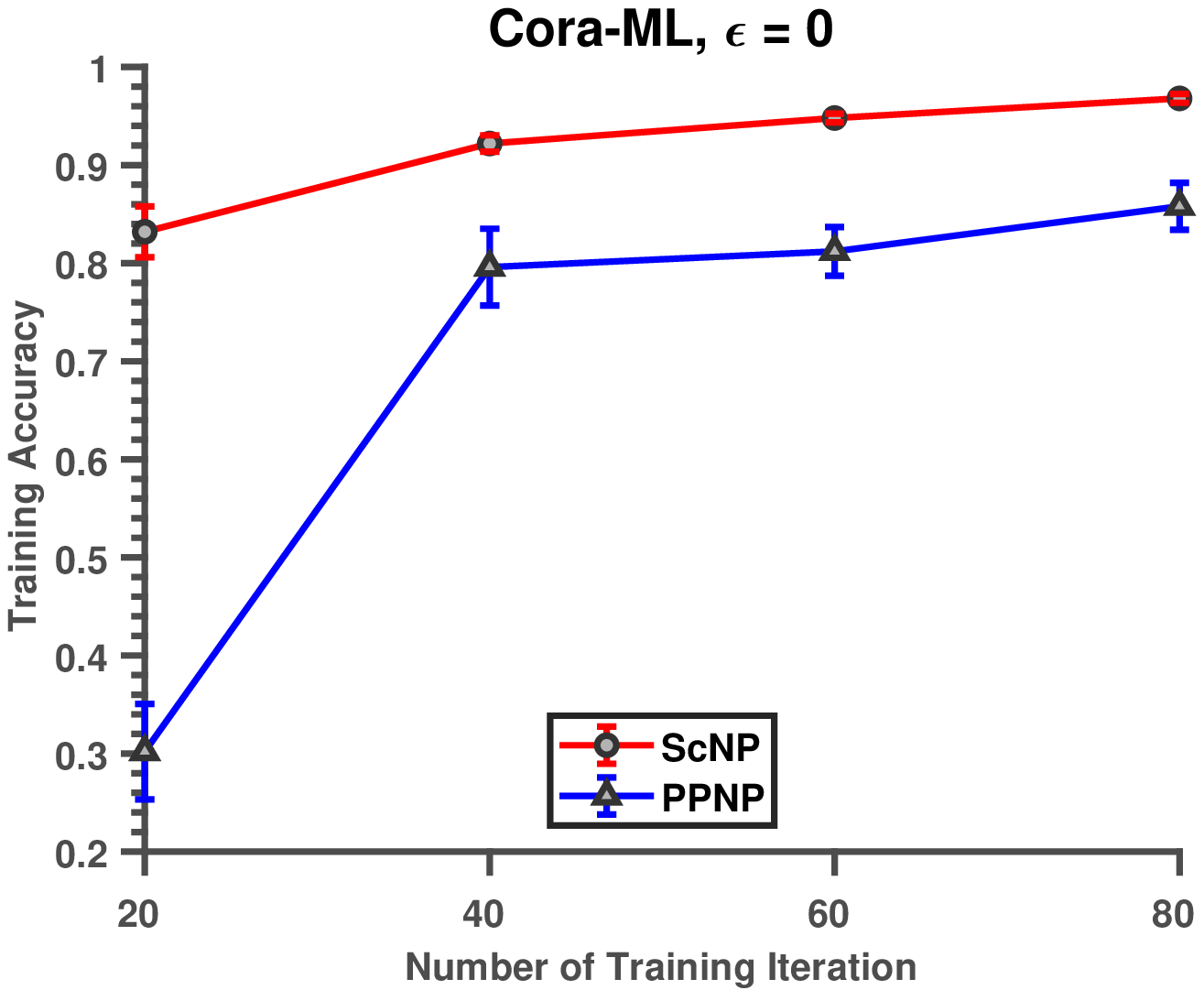}}\quad
   \subfigure[]{\label{fig:Result2b}
     \includegraphics[scale=0.5]{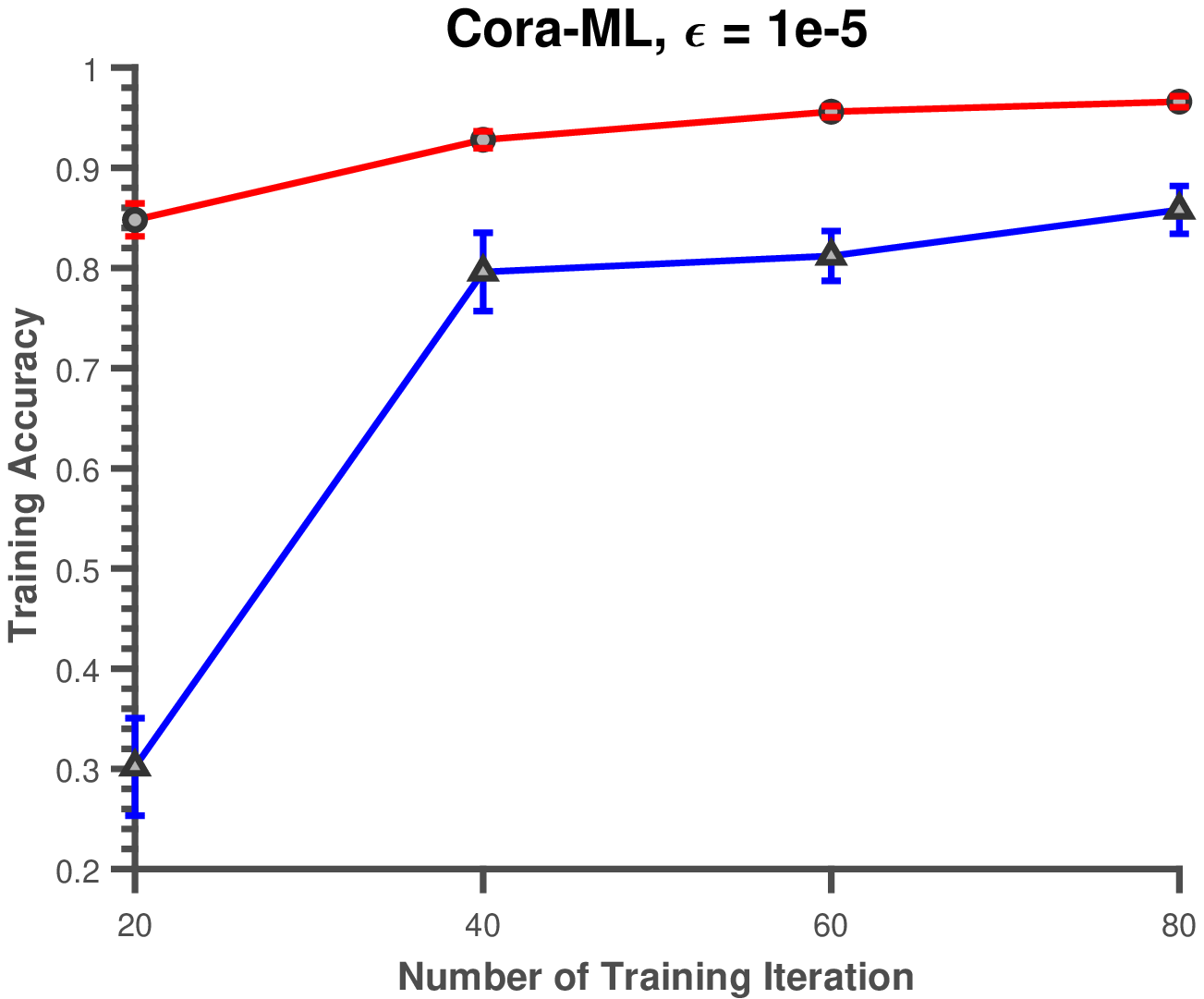}}\quad
    \subfigure[]{\label{fig:Result2c}
     \includegraphics[scale=0.5]{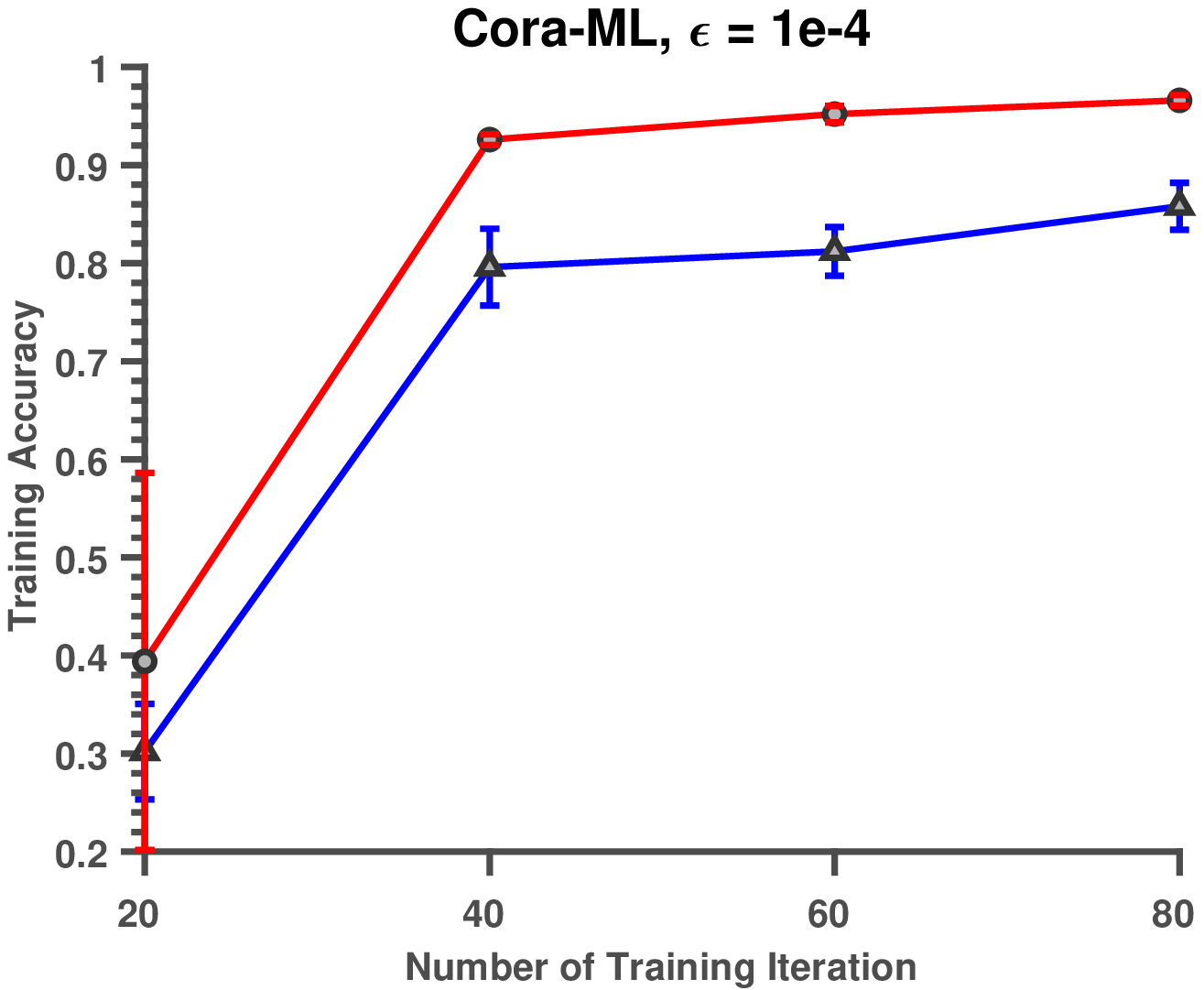}}\quad
    \subfigure[]{\label{fig:Result2d}
     \includegraphics[scale=0.5]{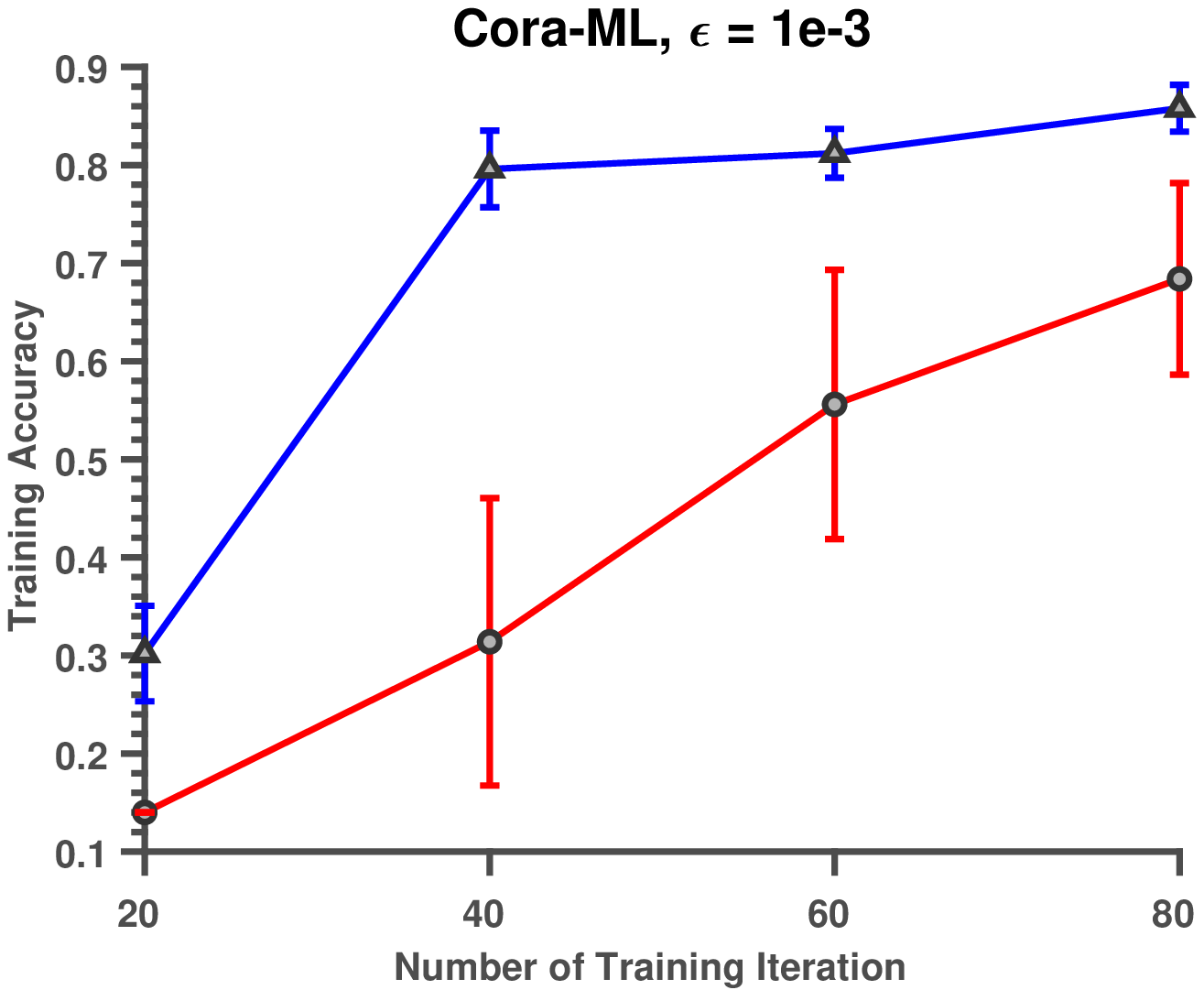}}\quad 
\end{center}
  
  \caption{\textbf{ Comparison of the performance of proposed message passing method  against that of the existing message passing algorithm  on the Cora dataset.} The plots show the mean and standard deviation of training accuracy of a GCN trained using {\sc ScNP} and {\sc PPNP}.  The performance of  is shown as a function of training {\em epochs}.}
  \label{fig:Performance}
\end{figure}

\begin{figure}[htbp]
  \begin{center}
  \subfigure[]{\label{fig:Result3a}
     \includegraphics[scale=0.5]{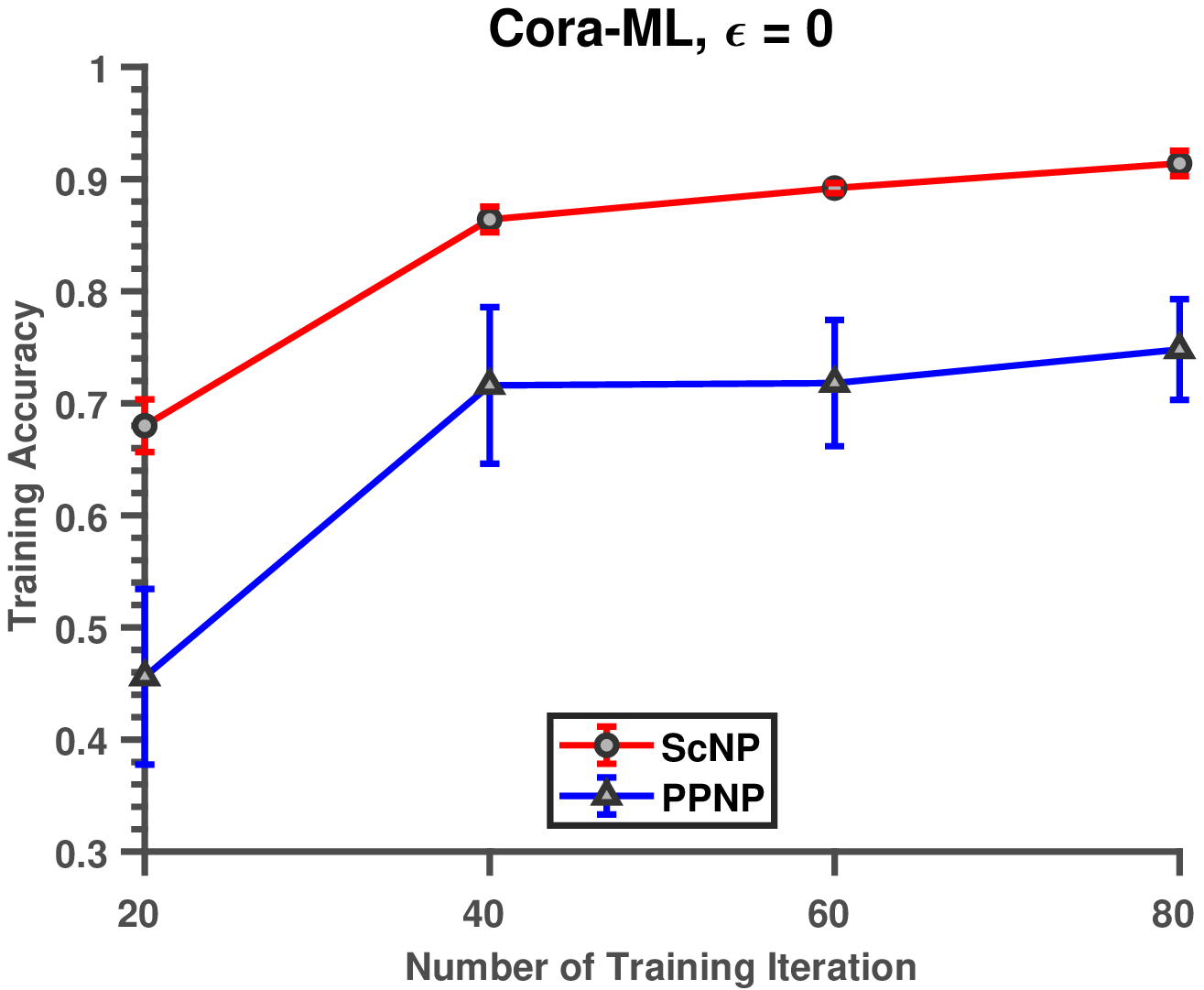}}\quad
   \subfigure[]{\label{fig:Result3b}
     \includegraphics[scale=0.5]{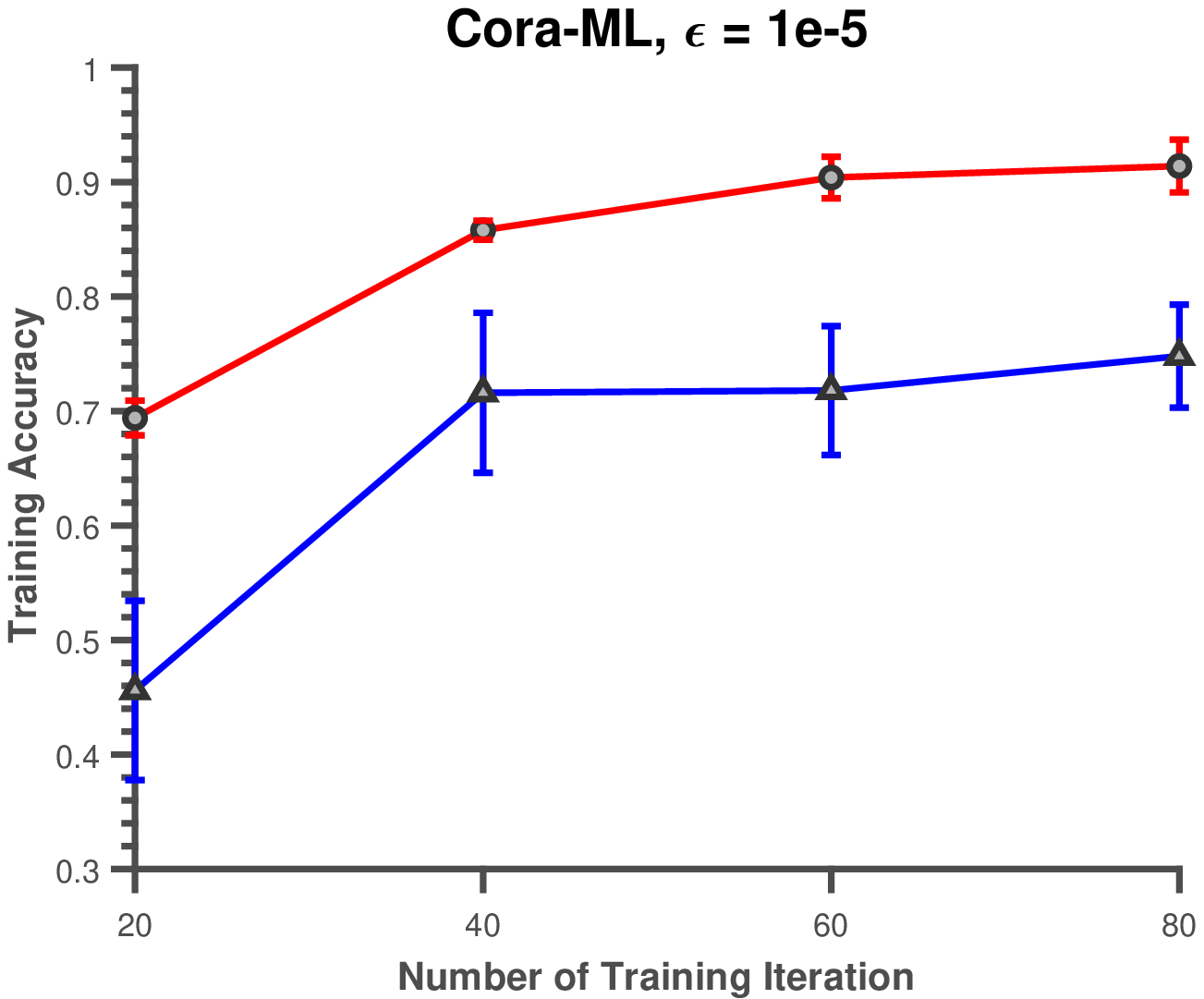}}\quad
    \subfigure[]{\label{fig:Result3c}
     \includegraphics[scale=0.5]{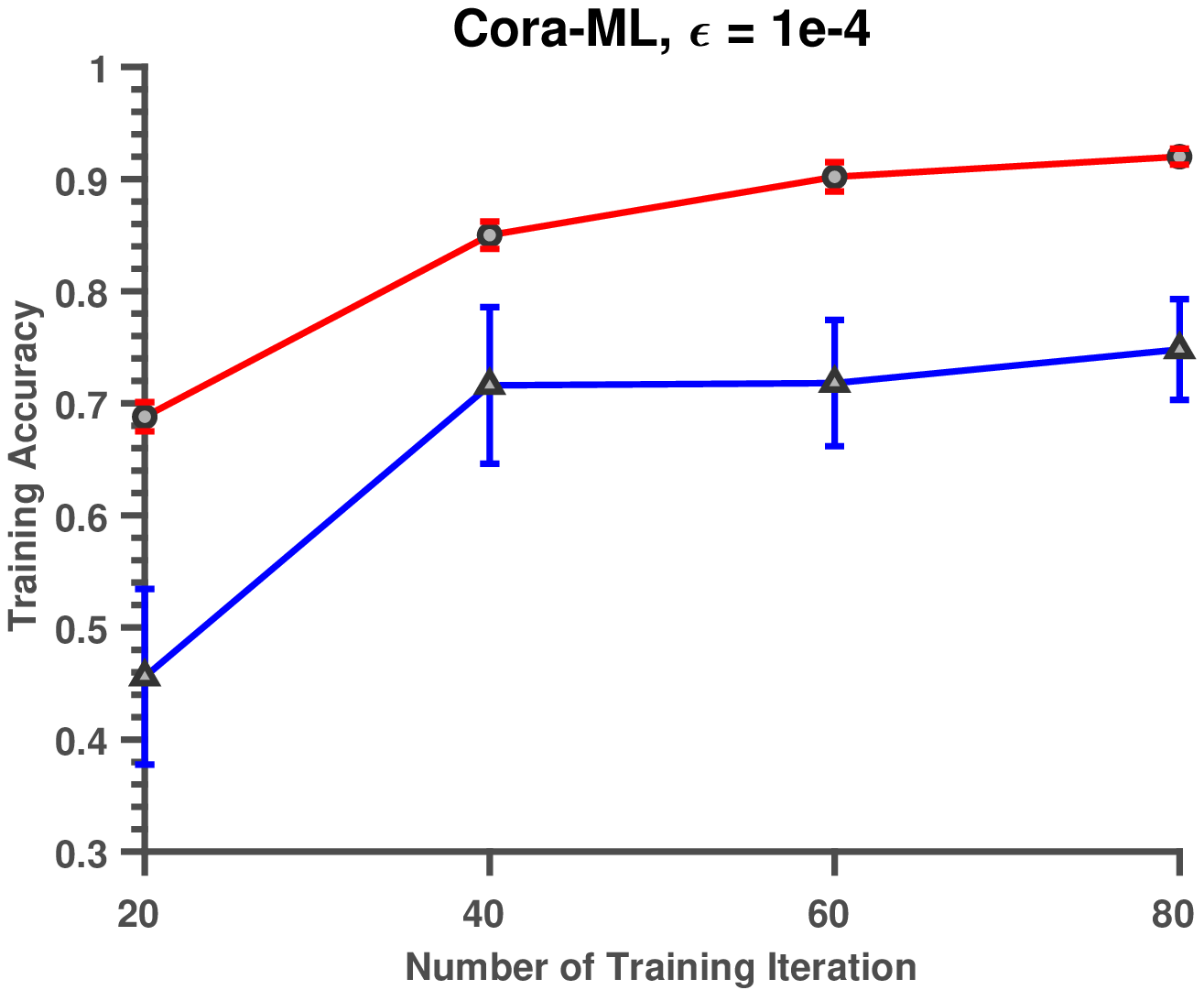}}\quad
    \subfigure[]{\label{fig:Result3d}
     \includegraphics[scale=0.5]{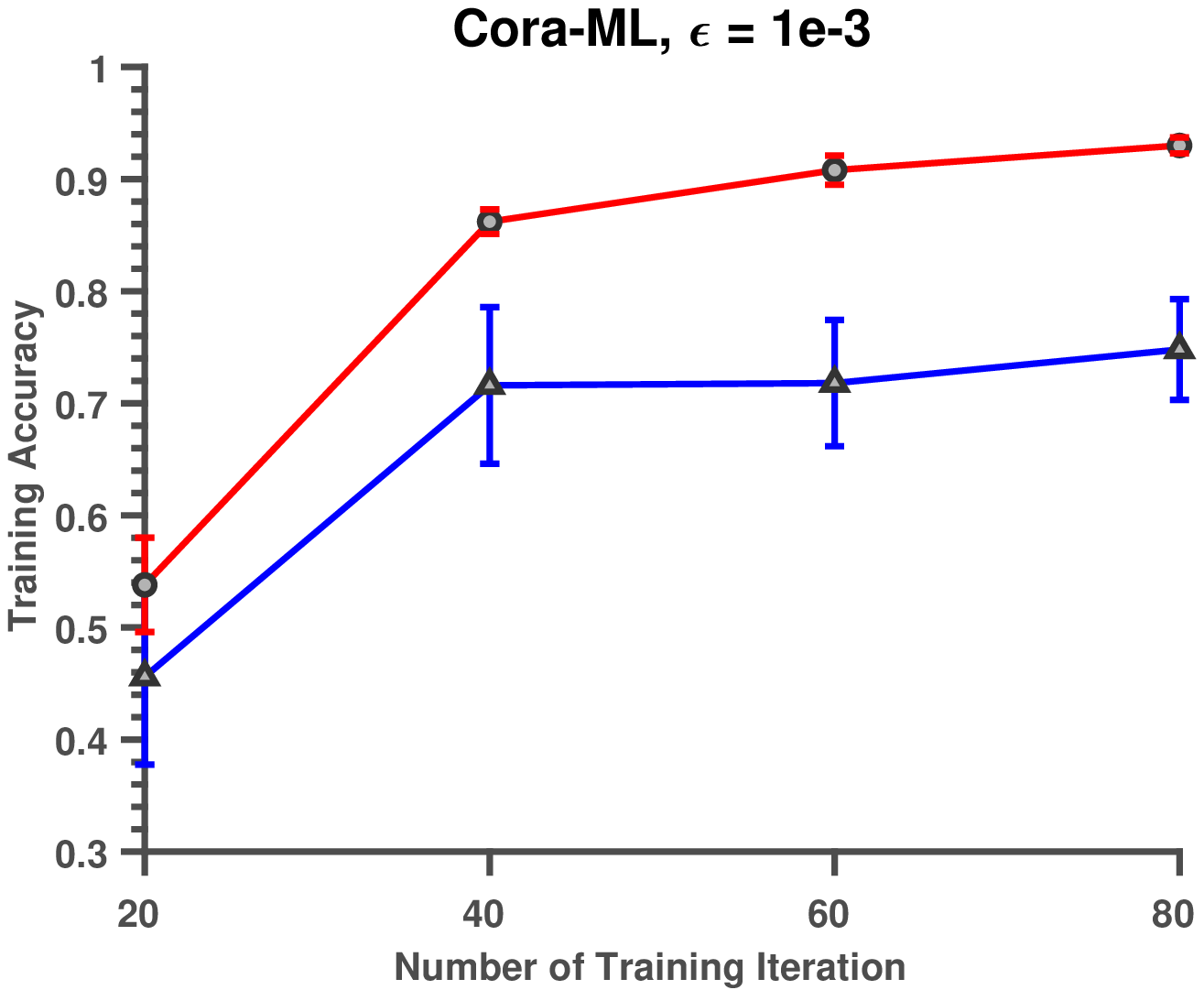}}\quad 
\end{center}
  
  \caption{\textbf{ Comparison of the performance of proposed message passing method  against that of the existing message passing algorithm  on the CiteSeer dataset.} The plots show the mean and standard deviation of training accuracy of a GCN trained using {\sc ScNP} and {\sc PPNP}.  The performance of  is shown as a function of training {\em epochs}.}
  \label{fig:PerformanceCite}
\end{figure}

\begin{figure}[htbp]
   \begin{center}
     \subfigure[]{\label{fig:CoraF31}
     \includegraphics[scale=0.45]{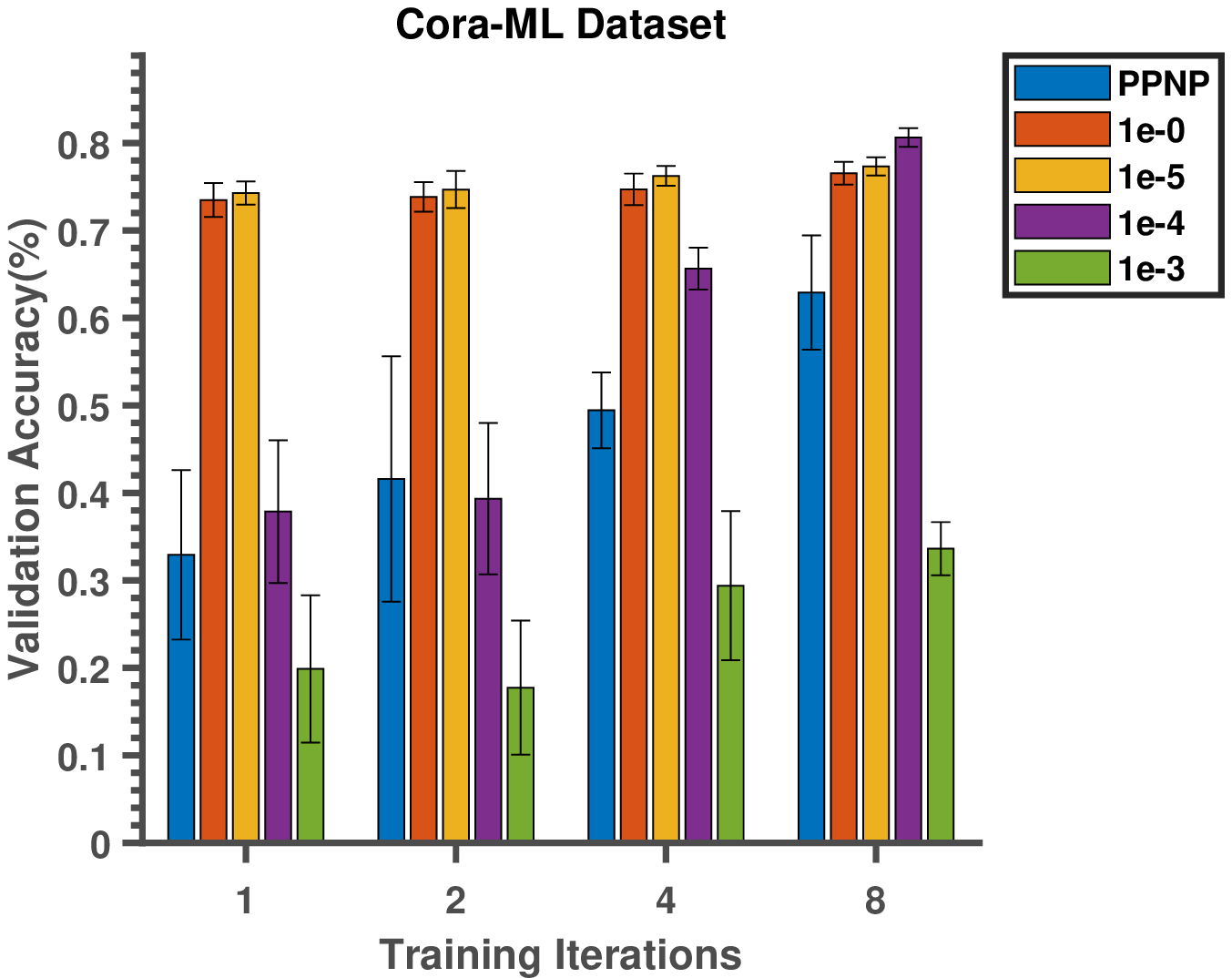}}\quad
     \subfigure[]{\label{fig:CoraF32}
     \includegraphics[scale=0.45]{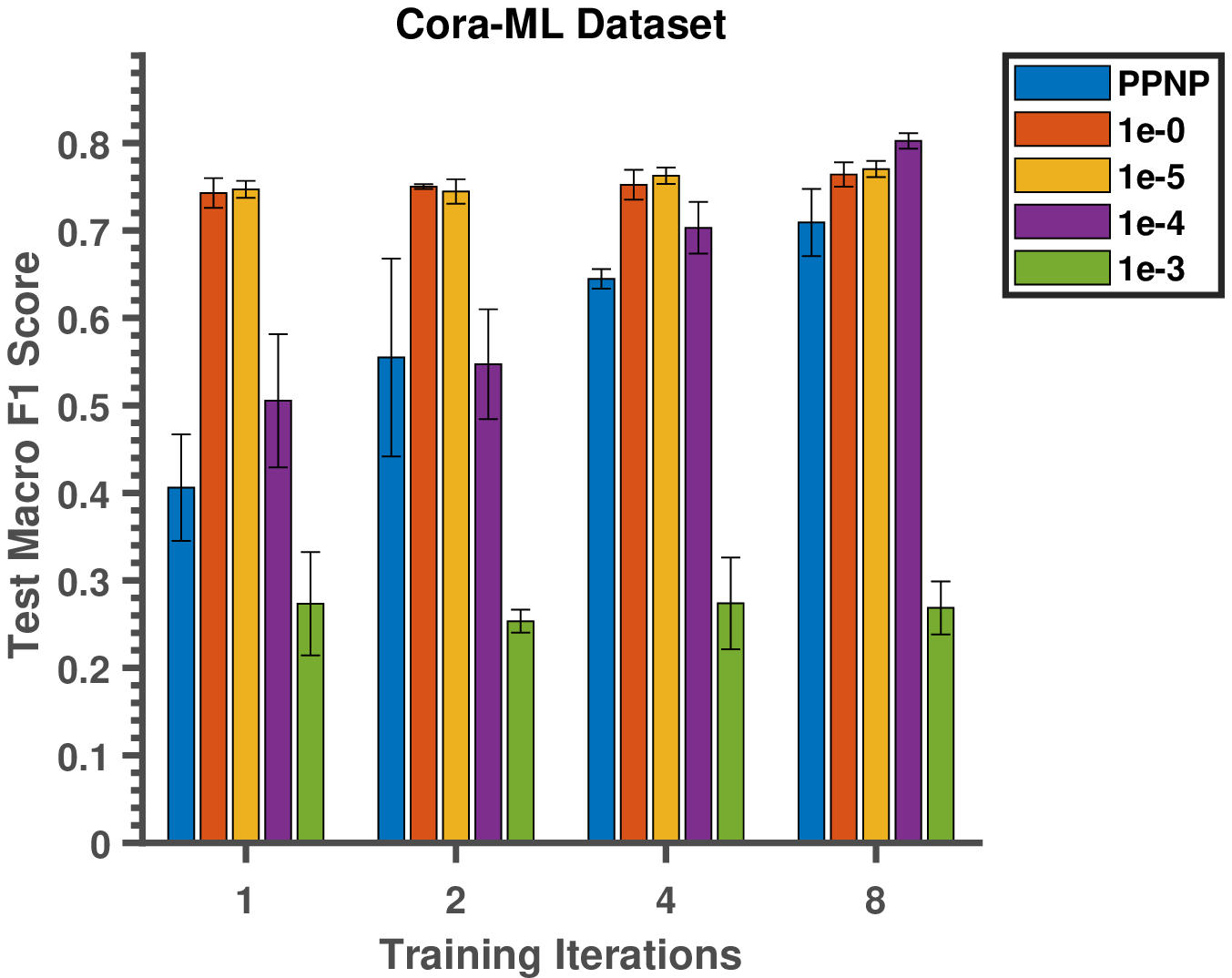}}\quad
      \subfigure[]{\label{fig:CiteF31}
      \includegraphics[scale=0.45]{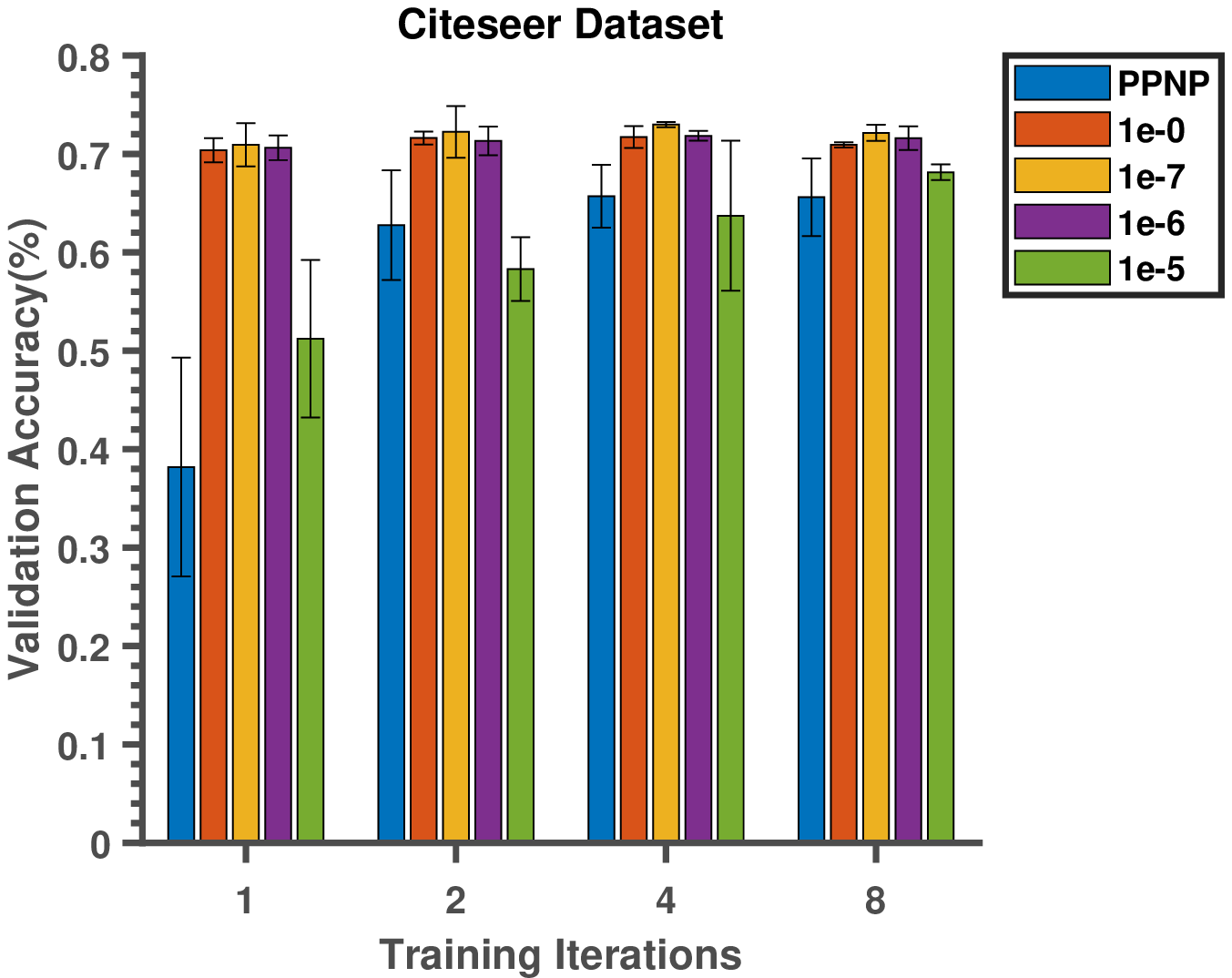}}\quad
      \subfigure[]{\label{fig:CiteF32}
      \includegraphics[scale=0.45]{CiteBarPlot1.eps}}\quad
      
   \end{center}
\caption{\textbf{The test accuracy and test Macro F1 of GCNs trained using our method and {\sc PPNP} with very limited training iterations.} In these experiments, we use various $\epsilon$ parameters across two datasets.}
   \label{fig:Performance3}
 \end{figure}
 
 In this section, we systematically evaluate the performances of proposed algorithmic framework in the context of the node classification problem. We start our discussion by describing the datasets and our experimental setting. 
Then, we analyze the performance of the algorithm as a function of the number of training iterations,{\em epochs}. 
We also compare the performance of our proposed methods against the best methods presented as {\em full PPNP},in~\cite{klicpera2018combining}.
We then investigate the performance of each algorithm as a function of the very limited training iterations.

\subsection{Datasets and Experimental Setup}
We test and compare the proposed methods on two sets of real-world collaboration networks: {\tt Cora} and {\tt CiteSeer} provided by~\cite{sen2008collective}. Details of these three networks are given on Table~\ref{datatable}. The datasets use a bag-of-words representation of the papers? abstracts as features., i.e., existence/non-existence of certain words are represented as $1/0$ values in this feature vector~\cite{AAAI18Li}.

For {\em PPNP}, we use the Python implementation provided by Klicpare {\em et al.}~\cite{klicpera2018combining}. 
We implement our {\em ScNP} in Matlab which is only used for offline computations, and use Python for online computation. For RWR based preprocessing we use both {\sc Chopper}~\cite{CoskunKdd} and {\sc I-Chopper}~\cite{CoskunVldb}. 

We assess the performance of the algorithm as a function training {\em epochs}. First, we fix the maximum training {\em epochs} to $80$ and presents the training validation accuracy with standard deviation in $10$ runs. Then, we evaluate the performance of the algorithm with very limited training {\em epochs}, namely for the values of $\{2^0, 2^1, 2^2,2^3\}$ and presents test validation accuracy and macro F-1 scores.

For the hyper-parameters of the GCNs, we follow Klicpare's~\cite{klicpera2018combining} parameter settings. Namely, we use a two layers
GCN with $h = 64$ hidden units. We apply $L_2$ regularization with $\lambda = 0.005$ on the weights of the first
layer and use dropout with dropout rate $d = 0.5$ on both layers and the adjacency matrix and $\alpha = 0.1$ for RWR. Finally, we report the mean accuracy of 10 runs for each dataset for the first experiment and 10 runs for the second experiment. Finally, for comparison we use full {\em PPNP} method as baseline method since it is the best method presented in~\cite{klicpera2018combining} and outperforms all opponents references (therein~\cite{klicpera2018combining}). All of the experiments are performed on a Dell PowerEdge T5100 server with two 2.4 GHz Intel Xeon E5530 processors and 32 GB of memory.

\subsection{Performance Evaluation}
We first compare the node classification performance of the our method and full {\em PPNP}~\cite{klicpera2018combining}
method using training accuracy, number of correct prediction divided by total number of prediction during the training phase of a GCN~\cite{AAAI18Li}, as the performance criterion. 
The results of this analysis for two datasets are shown in Figure~\ref{fig:Performance} and~\ref{fig:PerformanceCite}. 
As seen in the figures, on two datasets, the GCN that uses propagation based on our proposed algorithm delivers the best performance.
To be specific, on the Cora dataset, the training accuracy of the GCN that uses $\Sigma$ matrix instead of full personalized PageRank matrix, $\Pi$, in {\em PPNP} drastically outperforms its opponent, showing the validity of our method. Here, using very conservative $\epsilon$ value causes degradation of the method as seen in Figure~\ref{fig:Result3d}, suggesting that the $\epsilon$ parameters should be adjusted carefully for Cora dataset. On CiteSeer dataset, our method drastically improves training accuracy across for all $\epsilon$ values. Overall, the analyses in Figure~\ref{fig:Performance} and~\ref{fig:PerformanceCite} show that our method is highly effective for lifting the heavy burden of necessary usage of many training iteration for GCNs which in turn is very useful to computationally decrease the training time.

We then investigate the performance of our propagation algorithm as a function very limited number of training {\em epochs} and evaluate the test accuracy and Macro F1 scores, which is simple arithmetic mean of our per-class F1-scores, of its performance. 
The result of this analysis are shown in Figure~\ref{fig:Performance3}. 
As seen in the figure, the test accuracy and Macro F1 scores provided by our method is largely improved {\em PPNP} accuracy with very limited number of training {\em epochs}. It is very impressive that our method even renders a GCN classify nodes with only 1 training {\em epoch} showing that in the propagation phase of a GCN usage of {\sc ScPN} instead of {\sc PPNP} is much more true approach.   
These results clearly demonstrate the effectiveness of topological similarity based propagation instead of in regular personalized PageRank approach, suggesting that this algorithm has great potential in rendering GCNs message passing phase useful even when training with very limited number of {\em epochs}. These results demonstrate that our method will be new pioneer aspect for message passing phase of GCNs.

%% file: Conclusion.tex
\section{Conclusions}
\label{sec:conclusion}
In this paper, we propose an algorithmic framework that utilizes both graph intrinsic topology and features to remedy the message passing phase of Graph Convolutional Networks. Testing our method on real-world datasets, we demonstrate that our approach is highly effective in node classification problem and can be used for any graph convolutions network tasks that require message passing, such as link prediction.

The work presented here has many feature directions including usage of different matrix in a GCN instead of a Laplacian matrix, iterative computation of our method using effective Krylov base methods, etc to name a few.